\documentclass[11pt]{article}

\usepackage[preprint]{acl}

\usepackage{times}
\usepackage{latexsym}

\usepackage[T1]{fontenc}

\usepackage[utf8]{inputenc}

\usepackage{microtype}

\usepackage{inconsolata}
\usepackage{amssymb}
\usepackage{booktabs}
\usepackage{enumitem}
\usepackage{amsmath}
\usepackage{graphicx}
\usepackage{xcolor}
\definecolor{artifactHeader}{HTML}{344054}
\definecolor{artifactTrigger}{HTML}{EEF4F8}
\definecolor{artifactContent}{HTML}{F1F5EF}
\definecolor{artifactRule}{HTML}{667085}
\usepackage{colortbl}
\definecolor{hl}{gray}{0.88}

\usepackage{listings}
\title{Artifact Distillation: Training-Free Transfer from Frontier to Small Language Models}

\author{
  Mustafa Safa Ozdayi \qquad Somin Wadhwa \\
  Core AI, Qualtrics \\
  \texttt{\{mozdayi,swadhwa\}@qualtrics.com}
}

\begin{document}
\maketitle
\begin{abstract}
Closing the gap between frontier and smaller language models is usually approached by
distillation: fine-tuning the small model (student) on
a frontier model's
(teacher)
outputs, which binds transfer to training compute and leaves the transferred knowledge largely opaque. We propose \textit{Artifact Distillation}: a teacher solves training instances and externalizes its problem-solving procedure into an artifact bank that a student queries at inference time, with no gradient updates.
Across four LiveBench tasks, a bank written by Claude Opus 4.5 lets Claude Haiku 4.5
outperform unaided Claude Sonnet 4.5 on spatial reasoning (0.707 vs.\ 0.647) at 60\% of its cost, and reach Sonnet 4.5's level on table joining at 89\% of its cost.
The same raw bank raises Qwen3.5-27B's pooled score from 0.730 to 0.800, showing that transfer is not confined to the teacher's model family.
\end{abstract}

\section{Introduction}
Narrowing the performance disparity between frontier large language models (LLMs) and their smaller, cheaper counterparts has historically been approached through knowledge distillation via supervised fine-tuning (SFT): a student model is trained on outputs generated by a stronger teacher,
embedding its competence into updated weights \cite{kim2016sequence,hsieh2023distilling,guo2025deepseek}.
SFT binds transfer to training: it consumes compute, often risks catastrophic forgetting \cite{kirkpatrick2017overcoming, biderman2024lora}, and
often makes transferred knowledge difficult to inspect or revise.

In this work, we evaluate an alternative: leave the student's weights untouched, and externalize teacher knowledge directly into an artifact bank that the student can query at inference time. We call this framework \textit{Artifact Distillation}, a training-free form of distillation that transfers operational knowledge through retrievable artifacts rather than parameter updates. Artifacts are the methods the teacher used when solving the problems in natural language, making the transferred knowledge examinable, revisable, and reusable. We make the following contributions:

 \begin{itemize}[nosep,leftmargin=*]
    \item We introduce Artifact Distillation, a training-free method that converts a teacher's problem-solving capabilities into a transparent,
    observable artifact bank and delivers it to the student.
    \item We identify conditions under which a small student with the bank can be a cheaper alternative to a larger model.
    \item We find that the teacher--student gap does \textit{not} predict transfer: an artifact helps when it applies across instances and supplies a step the student is \textit{missing}.
\end{itemize}

\begin{figure*}%
  \centering
  \includegraphics[width=\textwidth]{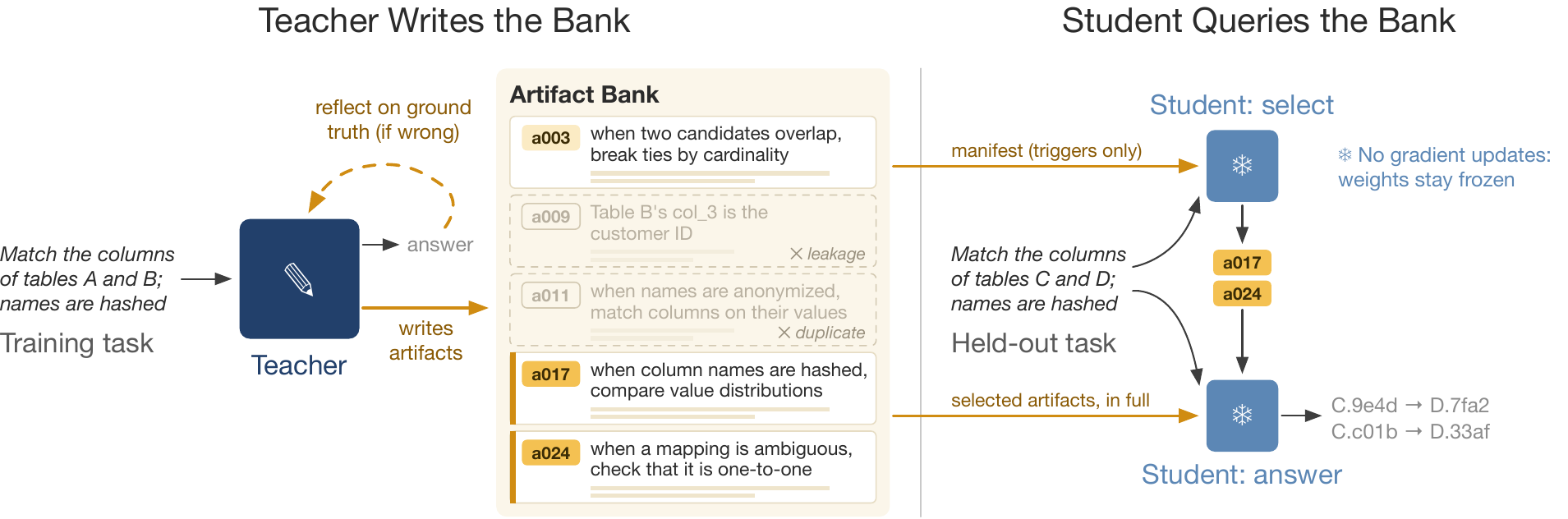}
  \caption{\textit{Artifact Distillation.} \textbf{Left:} the teacher solves training tasks and writes artifacts; when it answers incorrectly, a reflection pass with the
  ground truth writes new artifacts that replace the first attempt's. A judge then removes leaked, duplicate, trivial, or non-actionable artifacts (grayed). \textbf{Right:} at
  inference, the frozen student first selects artifact IDs from a manifest of triggers alone, then answers in a fresh call
  that receives only the selected artifacts in full.}
  \label{fig:main}
  \end{figure*}

\section{Method}
\label{sec:method}

Given a teacher $T$, a student $S$, and a task with a training split $D_{\mathrm{train}}$ and a held-out evaluation split $D_{\mathrm{eval}}$, the teacher builds from $D_{\mathrm{train}}$ a set of artifacts, which we refer to as the \textit{artifact bank}. We measure transfer as the change in the student's accuracy on $D_{\mathrm{eval}}$ when the bank is available, relative to the same student without it.

\paragraph{Teacher.}
For each training instance, $T$ first solves the problem and then writes one or more artifacts. An artifact consists of (i) a \textbf{unique identifier}, (ii) a \textbf{trigger} stating when it applies, and (iii) a free-text \textbf{content}. When the teacher's initial answer is not fully correct, a reflection call shows it the ground truth and asks for new artifacts, which replace those from the first attempt since those were written against flawed reasoning.

\begin{figure}[!ht]
\centering
\footnotesize
\setlength{\fboxsep}{5pt}

\colorbox{artifactHeader}{%
  \parbox{\dimexpr\columnwidth-2\fboxsep\relax}{%
  \color{white}\sffamily\bfseries
  ARTIFACT \texttt{a1}%
  \hfill\normalfont\scriptsize TABLEJOIN}}\\[-0.4pt]

\colorbox{artifactTrigger}{%
  \parbox{\dimexpr\columnwidth-2\fboxsep\relax}{%
  \textcolor{artifactRule}{%
    \sffamily\scriptsize\bfseries
    TRIGGER \hfill VISIBLE DURING SELECTION}\\[-1pt]
  When comparing columns across tables with obfuscated names,
  focus on the DATA TYPE and VALUE PATTERNS rather than trying
  to match row counts or positions.}}\\[-0.4pt]

\colorbox{artifactContent}{%
  \parbox{\dimexpr\columnwidth-2\fboxsep\relax}{%
  \textcolor{artifactRule}{%
    \sffamily\scriptsize\bfseries
    CONTENT \hfill REVEALED AFTER SELECTION}\\[-1pt]
  Categorical columns with a small set of distinct values
  (like ``spam''/``ham'', ``yes''/``no'', status codes) are
  often the easiest to match. Look for columns that share the
  exact same set of unique values.}}

\caption{An artifact from the \texttt{tablejoin} bank. Selection exposes only its identifier and trigger;
answering reveals the content after the student selects
\texttt{a1}. Appendix~\ref{app:tasks} gives the source problem.}
\label{fig:artifact}
\end{figure}

\paragraph{Bank cleaning.}
An LLM judge reads the whole bank in one call, so it can compare artifacts, and rejects those it considers duplicates, instance-specific, trivial, or not actionable.

\paragraph{Student.}
$S$ uses the bank in two stages. In the \textit{selection} stage, it sees the problem and a manifest of each artifact's identifier and trigger, and returns the identifiers it judges useful, or \texttt{NONE}. In the \textit{answer} stage, a fresh call receives the problem and the selected artifacts' contents.

\section{Experimental Setup}
\label{sec:experimental-setup}

\paragraph{Tasks.}
We select tasks with a measurable teacher--student gap. We
draw from LiveBench
\cite{white2025livebenchchallengingcontaminationlimitedllm},
restricted to instances not retired as of August 2026, and
screen six tasks from four of its categories
(Appendix~\ref{app:tasks}) using the unaided accuracy gap
$g = \mathrm{Acc}_{\text{Opus 4.5}} -
\mathrm{Acc}_{\text{Haiku 4.5}}$. From each category, we
retain the task with the largest significant gap. This yields \texttt{connections} (language;
$g=0.240$), which asks the model to sort 12--16 words into
hidden groups of four; \texttt{math\_comp} (math;
$g=0.386$), which contains competition mathematics
problems; \texttt{spatial} (reasoning; $g=0.233$), which
asks the model to count pieces or shapes produced by
cutting or arranging described solids; and
\texttt{tablejoin} (data analysis; $g=0.299$), which asks
the model to map columns between two tables when the column
names in one are replaced by opaque hashes, scored by F1.

\paragraph{Splits.}
We split each task's instances roughly 40/60 into training
and evaluation, stratified by problem subtype, with
near-duplicates kept on one side; Appendix~\ref{app:tasks}
gives pool sizes and bank-building costs.

\paragraph{Models.}
The teacher is Opus 4.5. We evaluate Haiku 4.5, Sonnet 4.5,
and the locally served Qwen3.5-27B
\cite{qwen2026qwen35} as students, and evaluate Opus
4.5 itself to test whether the teacher benefits from its
own bank. Opus 4.5 also serves as the cleaning
judge.\footnote{All models run in non-thinking mode at
temperature 1.} Appendix~\ref{app:prompts} gives all prompts.

\begin{table*}[ht]
\caption{Accuracy (mean of five runs) and inference cost per task. \$: inference cost in dollars per 1{,}000 instances; Qwen3.5-27B runs locally, so its cost is unreported. Pooled columns pair all 119 evaluation instances; for the raw and clean banks, $t$ is the paired $t$ statistic (mean over standard error) of the per-instance accuracy difference from the same model without a bank, and $|t|>2$ corresponds to two-sided significance at the 95\% level. Shaded: Haiku 4.5 with the raw bank vs.\ unaided Sonnet 4.5.}
\centering
\small
\setlength{\tabcolsep}{3.5pt}
\begin{tabular}{ll rr@{\hskip 9pt}rr@{\hskip 9pt}rr@{\hskip 9pt}rr@{\hskip 9pt}rrr}
\toprule
& & \multicolumn{2}{c}{\texttt{connections}} & \multicolumn{2}{c}{\texttt{math\_comp}} & \multicolumn{2}{c}{\texttt{spatial}} & \multicolumn{2}{c}{\texttt{tablejoin}} & \multicolumn{3}{c}{Pooled} \\
\cmidrule(lr){3-4}\cmidrule(lr){5-6}\cmidrule(lr){7-8}\cmidrule(lr){9-10}\cmidrule(lr){11-13}
Model & Bank & Acc. & \$ & Acc. & \$ & Acc. & \$ & Acc. & \$ & Acc. & $t$ & \$ \\
\midrule
Haiku  & --    & .676 & 4.6 & .559 & 6.6 & .587 & 3.9 & .498 & 2.6 & .580 &  & 4.4 \\
       & raw   & .666 & 7.4 & .552 & 7.6 & \cellcolor{hl}.707 & \cellcolor{hl}6.7 & \cellcolor{hl}.675 & \cellcolor{hl}8.9 & .651 & 3.9 & 7.7 \\
       & clean & .654 & 6.9 & .566 & 8.2 & .640 & 6.2 & .615 & 8.1 & .619 & 2.6 & 7.3 \\
\midrule
Sonnet & --    & .877 & 12.3 & .586 & 21.1 & \cellcolor{hl}.647 & \cellcolor{hl}11.2 & \cellcolor{hl}.698 & \cellcolor{hl}10.0 & .703 &  & 13.6 \\
       & raw   & .893 & 21.6 & .538 & 26.8 & .720 & 18.9 & .733 & 24.3 & .722 & 1.1 & 22.9 \\
       & clean & .899 & 20.8 & .559 & 25.5 & .733 & 18.1 & .739 & 23.1 & .734 & 1.9 & 21.9 \\
\midrule
Qwen   & --    & .772 & -- & .860 & -- & .882 & -- & .408 & -- & .730 &  & -- \\
       & raw   & .778 & -- & .882 & -- & .886 & -- & .671 & -- & .800 & 2.3 & -- \\
       & clean & .805 & -- & .882 & -- & .914 & -- & .686 & -- & .818 & 2.8 & -- \\
\midrule
Opus   & --    & .916 & 21.0 & .945 & 27.5 & .820 & 21.6 & .797 & 21.5 & .869 &  & 22.9 \\
       & raw   & .897 & 35.4 & .966 & 34.7 & .900 & 32.0 & .819 & 44.1 & .895 & 2.0 & 36.6 \\
       & clean & .914 & 34.2 & .938 & 33.7 & .827 & 30.5 & .821 & 41.2 & .874 & 0.4 & 34.9 \\
\bottomrule
\end{tabular}
\label{tab:main-results}
\end{table*}

\paragraph{Metrics.}
Accuracy is each task's official LiveBench grader score.
For the Claude models, cost is the per-instance inference
cost at Anthropic's list prices,\footnote{\url{https://docs.claude.com/en/docs/about-claude/pricing},
accessed August 2026.} summed over the student's full
pipeline of artifact selection and answer generation. We run each evaluation instance five times per
configuration and average its score.

\section{Results}
\label{sec:results}

The raw bank improves Haiku
4.5 by $+0.120$ on \texttt{spatial} and $+0.177$ on
\texttt{tablejoin}, while neither \texttt{connections} nor
\texttt{math\_comp} improves. This lets Haiku 4.5 outscore
unaided Sonnet 4.5 on \texttt{spatial} (0.707 vs.\ 0.647) at
60\% of its cost and reach Sonnet 4.5's level on
\texttt{tablejoin} (0.675 vs.\ 0.698) at 89\% of its cost. Pooled
raw-bank gains are $+0.071$ for Haiku 4.5 and $+0.070$ for
Qwen3.5-27B, but only $+0.020$ for Sonnet 4.5 and $+0.026$
for Opus 4.5.

\begin{table*}[t]
\caption{One of Haiku 4.5's most-selected raw-bank artifacts
on each task, shown with the training problem from which the
teacher wrote it (abridged; full text in
Appendix~\ref{app:tasks}). Sel.: share of evaluation runs on
which Haiku 4.5 selects the artifact. Gap: Opus 4.5 minus
Haiku 4.5 accuracy without artifacts. Rec.: share of that gap
recovered by the raw bank.}
\centering
\footnotesize
\setlength{\tabcolsep}{3.5pt}
\begin{tabular}{l p{0.29\textwidth} p{0.39\textwidth} rrr}
\toprule
Task & Training problem (abridged) & Artifact written from it
     & Sel. & Gap & Rec. \\
\midrule
\texttt{spatial}
& Cut regular heptagon along AC, AE, BG; how many pieces
  are quadrilaterals?
& \textit{When determining the shape of each piece:}
  count the vertices bounding each region, including interior
  intersection points of cuts%
& 53\% & .233 & 52\% \\

\texttt{tablejoin}
& Map Table A (spam/ham)
  to Table B's hashed columns.
& \textit{When column names are obfuscated:}
  match on data type and value patterns; categorical columns
  with a few distinct values (spam/ham) are easiest.
& 97\% & .299 & 59\% \\

\texttt{connections}
& Split 12 words, inc. \textit{bath, powder, rest, throne},
  into three groups of four.
& \textit{When words seem unrelated:}
  check if they precede/follow a common hidden
  word (eg, \underline{\hspace{1.2em}}room).
& 68\% & .240 & 0\% \\

\texttt{math\_comp}
& A triangle inscribed in a circle; find a chord length (AIME).
& \textit{Circumcircle problems:}
  Align one side with the $x$-axis and solve the circle's linear system.
& 10\% & .386 & 0\% \\
\bottomrule
\end{tabular}

\label{tab:when}
\end{table*}

\subsection{When Does Transfer Help?}
\label{sec:when}

We observe that the gap itself does not predict transfer. Haiku 4.5 trails
Opus 4.5 by nearly the same margin on \texttt{connections}
and \texttt{spatial}, yet the bank recovers half of the
\texttt{spatial} gap and none of the \texttt{connections}
gap. On \texttt{math\_comp}, which has the largest gap,
nothing transfers. We examine the artifact Haiku 4.5 selects
most often on each task to explain this split
(Table~\ref{tab:when}; Appendix~\ref{app:artifacts} gives the
top four per task in full).

The successful tasks share two properties. On
\texttt{spatial} and \texttt{tablejoin}, the most-selected
artifacts prescribe procedures that apply across instances
and that Haiku 4.5 skips when unaided. With the
\texttt{spatial} bank, the share of answers that trace each
region's boundary doubles from 23\% to 46\%. On
\texttt{tablejoin}, unaided Haiku 4.5 returns 2.2 column
mappings per problem where the gold answer has 3.2; with the
bank, it returns 3.2, raising recall from .47 to .70 and
precision from .61 to .68.

The failures each violate one condition. On
\texttt{connections}, the top artifact applies broadly but
supplies no new step: 87\% of unaided answers already test
for a shared hidden word. The bank raises secondary-sense
checks from 6\% to 21\% without improving accuracy, while
its remaining category lists help only when a category
recurs. On \texttt{math\_comp}, artifacts supply new but
narrow techniques; the selector returns \texttt{NONE} on
75\% of runs because few evaluation problems require
them. Transfer therefore requires both recurrence and novelty. Overall, to be useful, an artifact must apply across instances and supply a step the student is missing.

\subsection{Does Cleaning the Bank Help?}
\label{sec:cleaning}

We find that cleaning removes redundancy, but that redundancy is
sometimes functional. Across the four banks, the judge rejects 74 of 243 artifacts (30\%), 58
as duplicates. Haiku 4.5's pooled gain falls from $+0.071$
to $+0.039$, driven by \texttt{tablejoin} ($-0.060$) and
\texttt{spatial} ($-0.067$); Opus 4.5 also loses on
\texttt{spatial} ($-0.073$). Sonnet 4.5 remains essentially
unchanged, while Qwen3.5-27B loses on no task and gains
$+0.018$ pooled.

The losses expose two patterns. On \texttt{tablejoin}, the judge rejects three of Haiku 4.5's four most-selected artifacts as duplicates of the fourth (Figure~\ref{fig:artifact}); all four advise matching columns by their values. With the raw bank, Haiku 4.5's selected artifacts give this advice 3.9 times per instance; with the clean bank, 0.9 times. In its place, Haiku 4.5 selects contradictory advice: one artifact says to count exact value matches between columns and take the pairs with the most, while two others say the tables hold different records, so mapped columns should not be expected to share values. Haiku 4.5 selects the first with at least one of the others on 92\% of clean-bank runs, against 31\% with the raw bank, and its precision falls from 0.68 to 0.62. On \texttt{spatial}, the judge keeps the terser member in 11 of 16 duplicate pairs: 38\% of rejected duplicates contain a concrete example, against 16\% of retained artifacts. Examples make an artifact longer, but they may be what lets a student carry out the procedure, which would explain the (individually non-significant) \texttt{spatial} losses of Haiku 4.5 and Opus 4.5. Whether cleaning helps thus depends on the model--task pair.

\section{Related Work}
\label{sec:related-work}

Artifact Distillation changes \textit{where} transferred competence
resides. Standard distillation moves competence into the
student's weights by matching the teacher's output
distributions \cite{hinton2015distilling}, sequences
\cite{kim2016sequence}, intermediate representations
\cite{jiao2020tinybert}, or rationales
\cite{hsieh2023distilling,wadhwa2024investigating}; Artifact
Distillation leaves the weights untouched. Unlike RAG
\cite{lewis2020retrieval} or few-shot demonstrations
\cite{brown2020language}, including retrieved ones
\cite{rubin2022learning}, we
retrieve teacher-written guidance, not passages or solved
examples. Routing and cascade methods
\cite{chen2023frugalgpt,ong2025routellm} decide which
existing model answers each query; we instead make the
cheaper model itself more accurate.

Teacher-written guidance is the closest precedent. TPD
\cite{wang2024tpd} and RICP \cite{sun2024retrieved} both
have a teacher derive principles from the student's own
errors, requiring the student to first run on labeled non-test data.
TPD folds these
principles into revised instructions and few-shot examples,
while RICP retrieves
insights from
the training mistakes nearest each test question. To our
knowledge, neither compares the resulting student against a
stronger model run unaided on accuracy or cost. AutoManual
\cite{chen2024automanual} shows that a manual can guide
smaller LLMs, but builds it through interaction with an
environment; ExpeL and AutoGuide
\cite{zhao2024expel,fu2024autoguide} extract guidance from
agent experience, while other methods learn from a model's
own mistakes \cite{zhang2024incontext} or experience
\cite{shinn2023reflexion,ouyang2026reasoningbank}. We differ
in three ways: unlike TPD and RICP, the teacher writes our
bank from its own solves without observing a student,
allowing one bank to serve several models, including the
teacher; the student selects from triggers alone and sees
an artifact's content only after choosing it; and we compare
the student--bank system with a stronger unaided model on
cost as well as accuracy, showing it can match that model at
lower cost.

\section{Conclusions}
\label{sec:conclusion}

Artifact Distillation transfers a teacher's problem-solving procedure to a frozen student through a bank of natural-language artifacts. A bank written by Opus 4.5 lets Haiku 4.5 outscore unaided Sonnet 4.5 on \texttt{spatial} at 60\% of its cost and reach its level on \texttt{tablejoin} at 89\%, and the same bank raises Qwen3.5-27B's pooled score from 0.730 to 0.800. We find that an artifact helps when it applies across instances and supplies a step the student is missing. Cleaning the bank for concision is not free, since the repetition and examples a judge removes can be what a weaker student relies on.

\section*{Limitations}
This work has several limitations. First, although our four tasks span different LiveBench categories, they all come from a single benchmark, so how transfer behaves on other benchmarks or real workloads remains to be seen. Second, we consider only single-turn tasks; longer-horizon, agentic tasks are not covered, and they may be where a bank that the model queries on demand matters the most. Third, we show transfer from a Claude teacher to Qwen3.5-27B, but a single open-weight student does not establish patterns of transfer across model families; testing non-Claude teachers and students, such as OpenAI models, is a natural next step we leave for future work.

\bibliography{anthology,custom}
\clearpage
\appendix

\section{Prompts}
\label{app:prompts}
All prompts are given below. Placeholders in braces are filled at call time: \texttt{\{task\_domain\}} and \texttt{\{answer\_format\}} name the LiveBench task and its required answer format; \texttt{\{manifest\}} lists each artifact's identifier and trigger, and \texttt{\{artifacts\}} is the selected artifacts' full content; \texttt{\{task\}} is the problem statement; \texttt{\{ground\_truth\}} is the reference answer, shown only to the teacher when reflecting.

\subsection{Teacher: Generation (system)}
\label{app:prompt-teacher-system}
\begin{lstlisting}
You are acting as a TEACHER model. You will be given tasks about {task_domain} to complete. A separate, weaker STUDENT model will later face DIFFERENT instances of tasks from the same domain, with no access to this instance or your solution to it -- only to whatever artifacts you write below. Your job is to solve this instance well AND leave the student something genuinely useful for instances it hasn't seen yet.

After completing the task, you MUST write reusable artifacts for the student -- something that would genuinely change what the student does, not an obvious observation they'd land on anyway. Something that will help them avoid mistakes.

CRITICAL -- artifacts must be DOMAIN-GENERAL, not specific to THIS instance:
- The student will face a DIFFERENT instance of the same task (different inputs, different
  specifics). Write only guidance that transfers to ANY such instance.
- NEVER reference specific entities, names, values, file paths, or a fixed solution sequence from this instance. Triggers and content must describe a RECURRING structural situation, not this instance's content.
- BAD (overfit): "This problem involves Alice, so treat her as the narrator."
  GOOD (general): "WHEN ordering narrative events, prefer placing a character's introduction
  before any pronoun reference to them, because referents must be established first."
- If something only works for this one instance, do not write it. Prefer fewer, broadly
  applicable artifacts over many narrow ones.

There is no fixed taxonomy -- write whatever text would actually help: prose notes, a checklist, a worked pattern, or anything else you judge useful. Use your own judgment about form, structure, and how many to write.

Format your response EXACTLY as:

{answer_format}

<artifacts>
NOTE: [artifact_id] | [one-line trigger]
[Whatever would help the student. No fixed template.]

NOTE: [artifact_id] | [one-line trigger]
[Whatever would help -- prose, a checklist. No fixed template.]
</artifacts>
\end{lstlisting}

\subsection{Teacher: Reflection on Error (user)}
\label{app:prompt-teacher-reflection}
\begin{lstlisting}
Your answer was incorrect. The correct answer is: {ground_truth}

Review your reasoning above and identify exactly where and why you went wrong. Then write new artifacts -- following the rules specified in your system prompt above -- that would prevent this mistake in the future. Remember -- artifacts must be DOMAIN-GENERAL: capture the transferable lesson, never this instance's specific entities, names, values, or fixed solution sequence. The student will face a different instance of the same task.
\end{lstlisting}

\subsection{Bank Cleaning Judge (system)}
\label{app:prompt-bank-cleanup}
\begin{lstlisting}
<role>
You are auditing a bank of reusable problem-solving artifacts that a teacher model wrote while completing training tasks. These artifacts will be handed to a smaller student model completing DIFFERENT instances of the same task type. Your job is to decide, for each artifact, whether it survives into the cleaned bank -- NOT to rewrite or restructure it.

Each artifact's content has no fixed structure -- prose notes, checklists, whatever the teacher judged useful. Do not impose structure an entry doesn't have.
</role>

<input_format>
A JSONL bank, one artifact per line, each with `artifact_id`, `trigger`, `content`.
</input_format>

<criteria>
Reject an artifact if:
- LEAKAGE: it names a specific entity, plot point, number, or fixed solution sequence from the training instance instead of a recurring structural situation.
- DUPLICATE: another artifact in the bank already captures the same trigger and lesson, however it's worded or structured. Keep the clearer/more complete one, reject the rest -- cite the kept artifact's id.
- TRIVIAL: the advice is generic problem-solving common sense with no non-obvious information (e.g. "read the problem carefully").
- NOT_ACTIONABLE: the trigger is too vague to fire, or the guidance doesn't tell the student what to actually do differently.

Keep an artifact if it is domain-general, non-redundant, and would change what a capable student actually does when the trigger condition is met.
</criteria>

<instructions>
Judge the bank as a whole, not one artifact at a time: group artifacts by topic/trigger, treat every pair or cluster that overlaps in trigger and lesson as a DUPLICATE candidate, and for each such cluster keep whichever member is clearest/most complete. A DUPLICATE call is a comparison across the bank, not a property of one artifact read in isolation. Judge leakage and triviality in the same pass.
</instructions>

<output_format>
Respond with exactly one line per input artifact, in the same order. Format exactly as:

<verdicts>
[artifact_id] | KEEP | [one-line reason]
[artifact_id] | REJECT | [LEAKAGE|DUPLICATE|TRIVIAL|NOT_ACTIONABLE] | [one-line reason]
</verdicts>
</output_format>
\end{lstlisting}

\subsection{Student: Baseline (system)}
\label{app:prompt-student-baseline}
\begin{lstlisting}
<role>
You are an expert problem solver. Solve the task given to you.
</role>

<output_format>
Format your response EXACTLY as:

{answer_format}
</output_format>
\end{lstlisting}

\subsection{Student: Selection (system)}
\label{app:prompt-student-select-system}
\begin{lstlisting}
<role>
You are an artifact retrieval component inside a larger system. You do NOT solve tasks -- a separate model does that. Your only job is to decide which artifacts from a fixed library are relevant to an incoming task.
</role>

<manifest>
The library manifest is below, one artifact per line, formatted as `artifact_id | trigger`. The trigger describes when that artifact applies. Artifact content is deliberately withheld -- decide from the trigger alone.

{manifest}
</manifest>

<rules>
- Do NOT solve the task, work through it, or reason about what its answer might be. Solving is not your job, and any solution text you produce is discarded unread.
- Match on the task's STRUCTURE (what kind of problem it is), not on its specific numbers or names.
- Select the SMALLEST set of artifacts that covers the task -- not every artifact that seems loosely related. For each candidate, ask: would the solving model actually do something differently because of this one? If not, leave it out. When two triggers describe the same situation in different words, pick only the one that fits best.
- Selecting most or all of the library is almost always wrong, even on a task that touches many topics. A well-matched task is usually covered by a handful of artifacts, sometimes none. If you catch yourself including one "just in case", drop it.
</rules>

<output_format>
Output ONLY artifact_ids, one per line, copied exactly from the manifest. No prose, no numbering, no explanation, no preamble.

If no trigger matches, output the single word NONE.
</output_format>
\end{lstlisting}

\subsection{Student: Selection (user)}
\label{app:prompt-student-select-user}
\begin{lstlisting}
<instructions>
Select artifacts for the task below. Do not solve the task.
</instructions>

<task>
{task}
</task>

<output_format>
Output the relevant artifact_ids now, one per line, or NONE.

Begin your reply with the first artifact_id. Write nothing before it -- no analysis, no preamble.
</output_format>
\end{lstlisting}

\subsection{Student: Answer (system)}
\label{app:prompt-student-answer}
\begin{lstlisting}
<role>
You are an expert problem solver. You have been given artifacts -- strategies distilled from completing similar tasks. Use them where they apply as you solve the task.
</role>

<artifacts>
{artifacts}
</artifacts>

<output_format>
Format your response EXACTLY as:

{answer_format}
</output_format>
\end{lstlisting}

\section{Tasks, Splits, and Costs}
\label{app:tasks}

\paragraph{Task screen.} We screened six LiveBench tasks (\texttt{connections}, \texttt{math\_comp}, \texttt{spatial}, \texttt{tablejoin}, \texttt{AMPS\_Hard}, \texttt{cta}) by the unaided Haiku 4.5--Opus 4.5 gap and kept, from each LiveBench category, the task with the largest significant gap, which excluded \texttt{AMPS\_Hard} (math) and \texttt{cta} (data analysis).

\paragraph{Example problems.} One training problem per task, verbatim from LiveBench except where marked: the four problems Table~\ref{tab:when} abridges, each the source of the artifact shown there. All four were solved correctly by the teacher on its first attempt, so none of these artifacts comes from reflection.

\begin{quote}\small
\textbf{\texttt{connections}} (subtype: 3 groups). You are given 12 words/phrases below. Find three groups of four items that share something in common. [\ldots{} four example groups and answer-format instructions elided \ldots] The items are: warped, powder, lick, trace, bath, shred, bent, ounce, twisted, rest, throne, gnarly.\\
\textit{Answer:} bent, gnarly, twisted, warped / lick, ounce, shred, trace / bath, powder, rest, throne.

\textbf{\texttt{math\_comp}} (AIME I 2024). Let $ABC$ be a triangle inscribed in circle $\omega$. Let the tangents to $\omega$ at $B$ and $C$ intersect at point $D$, and let $\overline{AD}$ intersect $\omega$ at $P$. If $AB=5$, $BC=9$, and $AC=10$, $AP$ can be written as the form $\frac{m}{n}$, where $m$ and $n$ are relatively prime integers. Find $m+n$. [\ldots{} answer-format instructions elided \ldots]\\
\textit{Answer:} 113.

\textbf{\texttt{spatial}} (subtype: polygon chords). Suppose I have a physical, solid, regular heptagon with vertices ABCDEFG, and I make three cuts through AC, AE, and BG. Of the resulting pieces, how many quadrilaterals are there? [\ldots{} answer-format instructions elided \ldots]\\
\textit{Answer:} 1.

\textbf{\texttt{tablejoin}} (subtype: 1 pair). Please create a valid join mapping between CSV Table A and CSV Table B. Each column in A maps to 0 or 1 columns in B. Return your response as a Python dictionary, formatted as \{col\_name\_in\_df\_a : col\_name\_in\_df\_b\}. [\ldots]\\
Table A: \texttt{email,label}; rows such as \texttt{Act now! Limited-tim,spam} and \texttt{Upgrade to our premi,ham} (10 rows).\\
Table B: \texttt{lG1K/C5s5Ww,t8DtGa8xUVw}; rows such as \texttt{ham,0} and \texttt{spam,0} (11 rows).\\
\textit{Answer:} \{\texttt{label}: \texttt{lG1K/C5s5Ww}\}.
\end{quote}

\paragraph{Splits.} Each pool is the full set of non-retired instances. Splits are stratified proportionally by subtype; \texttt{math\_comp} is additionally stratified by solution topic, and near-duplicate instances (spatial instances sharing a configuration, tablejoin instances drawn from the same source dataset) are kept on one side.

\begin{table}[!ht]
\centering
\small
\setlength{\tabcolsep}{3pt}
\caption{Split sizes, the teacher's accuracy on the training half, bank size before $\rightarrow$ after cleaning, and the teacher's cost in dollars to build the raw bank (solving the training half and writing the artifacts; the cleaning call is not included).}
\label{tab:splits}
\begin{tabular}{lrrrrcr}
\toprule
Task & Pool & Train & Eval & Teacher & Artifacts & \$ \\
\midrule
\texttt{connections} & 50 & 20 & 30 & .79  & 69 $\rightarrow$ 52 & 0.82 \\
\texttt{math\_comp}  & 46 & 17 & 29 & 1.00 & 38 $\rightarrow$ 29 & 0.57 \\
\texttt{spatial}     & 50 & 20 & 30 & .85  & 59 $\rightarrow$ 38 & 0.75 \\
\texttt{tablejoin}   & 50 & 20 & 30 & .68  & 77 $\rightarrow$ 50 & 1.18 \\
\bottomrule
\end{tabular}
\end{table}

\paragraph{Costs.} Per-task inference costs are in Table~\ref{tab:main-results}. Building a bank costs \$0.57--\$1.18 in teacher calls (Table~\ref{tab:splits}). For \texttt{spatial}, Haiku 4.5 with the raw bank costs \$0.0067 per instance against \$0.0112 for unaided Sonnet 4.5, so the \$0.75 bank pays for itself after about 170 queries.

\section{Most-Selected Artifacts}
\label{app:artifacts}

We list, in full, the four raw-bank artifacts Haiku 4.5 selects most often on each task, with the share of evaluation runs on which each is selected (in parentheses); the trigger is in italics and followed by the content. We do not report a per-artifact effect on accuracy: the score when an artifact is selected is confounded with which instances it is selected on.

\small
\paragraph{\texttt{connections}.}
\begin{itemize}[nosep,leftmargin=*]
\item[] \textbf{a20} (68\%) \textit{When you see words that seem unrelated, check if they can all precede or follow a common word} \\ Many puzzles use compound words or phrases where all items combine with the same hidden word. Test common words like ``room,'' ``house,'' ``man,'' ``light,'' ``water,'' etc. as potential connectors before or after each item.
\item[] \textbf{a51} (54\%) \textit{Common words often have overlooked secondary meanings} \\ Words with everyday meanings (like ``treasure'' = valuable object) often have secondary uses (like ``treasure'' = term of endearment). When a word seems to fit one category, actively ask: ``What OTHER category could this word fit?'' especially for nouns that might also be pet names, verbs, or parts of compound phrases.
\item[] \textbf{a53} (50\%) \textit{When words seem to form compounds with a common word, check if they're actually synonyms FOR that thing} \\ A critical distinction: ``credit, charge, plastic, card'' are all SLANG TERMS FOR a credit/payment card (they refer to the same object), NOT words that combine with ``card'' to form compounds. Before assuming a ``\_\_\_ X'' or ``X \_\_\_'' compound pattern, ask: could these words all BE synonyms for the same concept? ``Plastic,'' ``card,'' ``credit,'' and ``charge'' are all ways people refer to payment cards.
\item[] \textbf{a57} (45\%) \textit{When a word fits multiple candidate categories, test all possible groupings before committing} \\ Words often have multiple meanings or associations. A word like ``perfect'' might seem to fit a ``pristine condition'' group (perfect condition) BUT also fit a structural pattern (perfect square). Before finalizing groups: \\ 1. For each word that seems to fit a category, actively search for OTHER categories it could fit \\ 2. Test whether removing that word from your initial group still leaves exactly 4 items \\ 3. If a word fits two groups, check which grouping allows ALL 16 words to form valid groups of 4 \\ 4. The puzzle has exactly one solution --- a word that ``works'' in one group may NEED to go elsewhere
\end{itemize}
\paragraph{\texttt{math\_comp}.}
\begin{itemize}[nosep,leftmargin=*]
\item[] \textbf{a34} (10\%) \textit{Coordinate geometry for triangle circumcircle problems} \\ Place one side on the x-axis for simpler calculations. The circumcircle equation x$^2$ + y$^2$ + Dx + Ey + F = 0 with three points gives a linear system. Tangent at point (x$_0$,y$_0$) is: x·x$_0$ + y·y$_0$ + D(x+x$_0$)/2 + E(y+y$_0$)/2 + F = 0.
\item[] \textbf{a14} (9\%) \textit{When points lie on a circle centered at the origin} \\ When multiple points lie on a circle centered at the origin, they are equidistant from the origin. Set up the equation by equating the squares of their distances: $x_1^2 + y_1^2 = x_2^2 + y_2^2$. This converts the geometric condition into an algebraic equation in the unknown parameter.
\item[] \textbf{a3} (6\%) \textit{When asked for a product/sum of variables} \\ You often don't need to find individual variable values. Setting up relationships and using identities may directly yield the requested product/sum, even when individual values require solving transcendental equations.
\item[] \textbf{a2} (3\%) \textit{When seeing $\log_a(b^c)$ in equations} \\ Use the power rule immediately: $\log_a(b^c) = c \cdot \log_a(b)$. This converts the problem into finding $\log_a(b)$, which often connects to simpler relationships.
\end{itemize}
\paragraph{\texttt{spatial}.}
\begin{itemize}[nosep,leftmargin=*]
\item[] \textbf{a46} (53\%) \textit{When determining the shape type of each resulting piece} \\ Count the number of distinct vertices/corners bounding each region. Vertices include: original polygon vertices AND interior intersection points of cuts. A region with 3 boundary vertices is a triangle, 4 is a quadrilateral, 5 is a pentagon, etc. Don't assume---trace each region's boundary completely.
\item[] \textbf{a30} (53\%) \textit{Systematic approach for cut problems} \\ When analyzing how cuts divide a shape, systematically enumerate which pairs of edges each cut can cross. For n parallel cuts through a triangle, consider all valid combinations of edge-pairs for each cut. Don't assume constraints (like ``must cross same edges'') without verification---draw specific counterexamples to test your assumptions.
\item[] \textbf{a58} (47\%) \textit{For a convex polygon with k internal intersection points from cuts, count regions by Euler's formula} \\ After making diagonal cuts in a convex polygon: count vertices (V = original vertices + intersection points), edges (E = original edges + cut segments, where each intersection splits segments), and use F = E - V + 2 (one face is exterior). Alternatively, trace boundaries carefully. Each time two cuts cross, they create additional regions.
\item[] \textbf{a56} (46\%) \textit{When cutting a convex polygon with diagonals, count intersection points and trace regions systematically} \\ When two diagonals of a convex polygon intersect inside, label the intersection point (e.g., P). Then systematically trace each closed region by following edges and cut segments. Each region's shape depends on how many boundary segments it has (3 = triangle, 4 = quadrilateral, etc.).
\end{itemize}
\paragraph{\texttt{tablejoin}.}
\begin{itemize}[nosep,leftmargin=*]
\item[] \textbf{a4} (100\%) \textit{When column names are hashed/obfuscated, your primary tool is examining the actual cell values.} \\ Create a mental inventory: What are the unique values in each column? What's the data type? What's the cardinality? Match columns that share these characteristics.
\item[] \textbf{a9} (99\%) \textit{When column names are opaque hashes, focus entirely on comparing the SET of unique values in each column} \\ To identify column mappings when target column names are meaningless: \\ 1. Extract unique values from each column in both tables \\ 2. Look for exact or near-exact overlap in value sets between columns \\ 3. Categorical columns with small, distinct value sets (like ``Weak/Strong'' or ``Yes/No'') are easiest to match \\ 4. Be careful not to confuse columns with similar data types but different semantics (e.g., two columns both containing strings)
\item[] \textbf{a49} (99\%) \textit{When column names are opaque hashes, focus entirely on value patterns} \\ Compare the actual data values between tables: look at data types (integers, floats, strings), value ranges, cardinality (how many unique values), and specific recurring values. Column names in Table B are meaningless hashes, so the ONLY signal comes from matching value distributions.
\item[] \textbf{a1} (97\%) \textit{When comparing columns across tables with obfuscated names, focus on the DATA TYPE and VALUE PATTERNS rather than trying to match row counts or positions.} \\ Categorical columns with a small set of distinct values (like ``spam''/``ham'', ``yes''/``no'', status codes) are often the easiest to match. Look for columns that share the exact same set of unique values.
\end{itemize}
\normalsize

\end{document}